\documentclass[runningheads]{llncs}
\usepackage[T1]{fontenc}

\usepackage{microtype}
\usepackage{graphicx}
\usepackage{subfigure}
\usepackage{booktabs} 

\usepackage{hyperref}
\usepackage{wrapfig, lipsum}
\usepackage[utf8]{inputenc}
\usepackage{microtype}
\usepackage{graphicx}
\usepackage{booktabs} 
\usepackage{url}            
\usepackage{nicefrac}       
\usepackage{xspace}
\usepackage{enumerate}
\usepackage{amsmath, mathtools, amsfonts, amssymb, dsfont, enumitem, bm, bbm, mathabx}
\usepackage{graphicx}
\usepackage[mathscr]{euscript}
\usepackage[capitalize,noabbrev]{cleveref}
\usepackage{adjustbox}
\usepackage{footmisc}
\usepackage[skip=5pt]{caption}
\usepackage[skip=3pt]{subcaption}

\usepackage{soul}
\usepackage[capitalize,noabbrev]{cleveref}
\usepackage{floatrow}

\usepackage{booktabs}
\usepackage{multirow}

\usepackage{amsmath}
\usepackage{amssymb}
\usepackage{mathtools}

\usepackage[capitalize,noabbrev]{cleveref}

\usepackage[textsize=tiny]{todonotes}

\usepackage{placeins}
\usepackage[utf8]{inputenc}
\usepackage{colortbl}
\usepackage{url}
\usepackage{amsmath} 
\usepackage{amssymb}  
\usepackage{graphicx}
\usepackage{hyperref}
\usepackage{mathtools}
\usepackage{xcolor}
\usepackage{bm}
\usepackage{graphicx}
\usepackage{siunitx}
\usepackage{tabularx}

\newcommand{\eps}{\varepsilon}

\renewcommand{\bar}{\widebar}

\newcommand{\R}{\mathbb{R}}

\newcommand{\N}{\mathcal{N}}

\newcommand{\E}{\operatorname{\mathbb{E}}}

\newcommand{\norm}[1]{\left\|#1 \right\|}

\usepackage{graphicx}
\begin{document}
\title{Constraint-Aware Diffusion Models for Trajectory Optimization}
%
%
\author{Anjian Li\inst{1}
\and Zihan Ding\inst{1}
\and Adji Bousso Dieng\inst{2,3}
\and Ryne Beeson\inst{4}}
\authorrunning{A. Li et al.}
%
\hypersetup{
    colorlinks=true,
    linkcolor=blue,
    urlcolor=blue,
    citecolor=blue,
    pdfborder={0 0 0}
}
\institute{Department of Electrical and Computer Engineering, Princeton University 
\and Department of Computer Science, Princeton University
\and \href{https://vertaix.princeton.edu/}{Vertaix}
\and Department of Mechanical and Aerospace Engineering, Princeton University 
\\
\email{\{anjianl, zihand, adji, ryne\}@princeton.edu}}
\maketitle              
\begin{abstract}
The diffusion model has shown success in generating high-quality and diverse solutions to trajectory optimization problems.
However, diffusion models with neural networks inevitably make prediction errors, which leads to constraint violations such as unmet goals or collisions.
This paper presents a novel constraint-aware diffusion model for trajectory optimization.
We introduce a novel hybrid loss function for training that minimizes the constraint violation of diffusion samples compared to the groundtruth while recovering the original data distribution. 
Our model is demonstrated on tabletop manipulation and two-car reach-avoid problems, outperforming traditional diffusion models in minimizing constraint violations while generating samples close to locally optimal solutions.

\end{abstract}

\section{Introduction}

Diffusion models \cite{sohl2015deep,ho2020denoising,song2020score} recently show promising performance on sampling high-quality data including images \cite{nichol2021improved,rombach2022high}, videos \cite{ho2022video}, robot trajectories\cite{janner2022planning,ajay2022conditional,chi2023diffusion}, etc.
In particular, 
with the ability to model complex data distribution, diffusion models can produce near-optimal and diverse trajectory candidates for efficient trajectory design that generalize to unseen environments \cite{li2024efficient}.
Those trajectories need to satisfy certain constraints that account for system dynamics, collision avoidance, goal-reaching, etc.

Unlike black-box optimization problems in biology, chemistry, or material science\cite{trabucco2022design,krishnamoorthy2023diffusion}, constraints for trajectory optimization often have analytical forms, e.g. the minimum distance between trajectory and obstacle needs to be above some safety threshold.
However, encoding such constraint information into the diffusion model has two major challenges: (1) The diffusion model learns the error between data in each diffusion step \cite{ho2020denoising} which makes it hard to directly evaluate the constraint violations.
(2) In the diffusion process, the data is gradually corrupted with increasing noise scales and inherently violates the constraints in the intermediate sampling steps.

In this paper, we propose a novel diffusion model that enhances constraint satisfaction while maintaining the high quality of the sampled solutions.
During training, we use a one-step reverse sampling to generate data samples with noise predictions from the neural network.
We introduce an additional loss function to evaluate constraint violations of these samples based on the groundtruth corrupted data.
We demonstrate the efficacy of our methods with two examples: tabletop manipulation and two-car reach-avoid problems.
Our models generate trajectories that warm-start the numerical trajectory optimization quickly, indicating that the samples are close to locally optimal solutions.
Additionally, our constraint-aware diffusion model shows fewer constraint violations compared to traditional unconstrained diffusion models.\looseness=-1

\section{Related Work}

\paragraph{Diffusion Models.}
Diffusion model \cite{sohl2015deep,ho2020denoising,song2020score} is a generative model with a forward diffusion process that gradually adds noises to the data and a learnable reverse process.
It can incorporate context information to achieve conditional generation with classifier guidance \cite{dhariwal2021diffusion} or classifier-free guidance \cite{ho2022classifier} methods.
Diffusion models have been successfully applied to generating image and video \cite{nichol2021improved,rombach2022high,ho2022video}, protein structure/molecule \cite{hoogeboom2022equivariant,ketata2023diffdock}, and robot controls \cite{janner2022planning,ajay2022conditional,chi2023diffusion,li2024efficient}.

\paragraph{Diffusion Model with Constraints.}
Constraint information has been used as a classifier \cite{botteghi2023trajectory,maze2023diffusion} to guide the diffusion sampling process or a diffusion kernel \cite{chang2023denoising}.
Graph-based constraints are also incorporated in diffusion models in combinatorial optimization \cite{sun2024difusco}.
Different constraints can also be composed for diffusion model \cite{yang2023compositional,power2023sampling}
In this paper, we aim to incorporate the constraint from a trajectory optimization problem through a novel loss function for training.
\section{Method}

In this section, we first introduce the trajectory optimization problem and the basics of the vanilla unconstrained diffusion model proposed in the previous literature \cite{ho2020denoising,ho2022classifier}.
Then we present our novel constraint-aware diffusion model in detail.

\subsection{Trajectory Optimization Problem}
In this paper, we aim to solve a trajectory optimization problem which can be formulated as a parameterized Nonlinear Program (NLP):
\begin{align} \label{eq: trajectory optimization prob}
    \mathcal{P}_{y} \coloneqq 
    \begin{cases}
    \underset{x}{\min} \quad &J(x;y)  \\
    s.t., \quad &g_{i}(x;y) \leq 0, \ i = 1, 2, ..., l \\
    \quad &h_{j}(x;y) = 0, \ j = 1, 2, ..., m  \\
    \end{cases}  
\end{align}
where $J \in \mathcal{C}^1(\R^n, \R^k; \R)$ is the objective function to minimize, e.g. the time to reach the goal or fuel expenditure.
$g_i \in \mathcal{C}^1(\R^n, \R^k; \R)$ and $h_j \in \mathcal{C}^1(\R^n, \R^k; \R)$ are both constraint functions including dynamics constraints, collision avoidance, goal-reaching, etc.
$x \in \R^n$ is the optimization variable such as controls.
$y \in \R^k$ represents the problem parameter that will vary as the task settings.
For example, $y$ could contain goal or obstacle positions, control limits, etc.

To generate training data for the diffusion models, we use Sparse Nonlinear OPTimizer (SNOPT) \cite{gill2005snopt}, a gradient-based numerical solver based on sequential quadratic programming (SQP) \cite{boggs1995sequential}, to solve the problems $\mathcal{P}_{y}$.
To create diverse problem instances, we uniformly sample $y$ within a reasonable range to obtain a collection of problems $\mathcal{P}_{y}$.
For each $\mathcal{P}_{y}$, the variable $x$ is uniformly sampled for the initial guesses from which the solver can obtain locally optimal thereby feasible solutions $x^*$.
Finally, the diffusion model is trained to learn $p(x^*|y)$.

\subsection{Unconstrained Diffusion Model}

We use the Denoising Diffusion Probabilistic Model (DDPM) \cite{ho2020denoising} as our unconstrained baseline with classifier-free guidance \cite{ho2022classifier}.
In this vanilla diffusion model, the forward sampling process is to gradually add Gaussian noise to the dataset with $K$ sampling steps $q(x_{k+1}|x_k) = \mathcal{N}(x_{k+1};\sqrt{1 - \beta_k}x_k, \beta_k I), 0 \leq k \leq K-1$.
$\beta_k$ is a variance schedule that increases from 0 to 1.
Thus we can obtain $x_k$ with a closed form given $x_0$ with $\alpha_k = 1 - \beta_k, \bar{\alpha}_k = \prod_{i=1}^K (1 -\beta_{i})$ and $\eps \sim \N(0, I)$:
\begin{align} \label{eq: x_k closed form}
    x_k = \sqrt{\bar{\alpha}_k} x_0 + \sqrt{1 - \bar{\alpha}_k} \eps,
\end{align}
At the final sampling step $K$, $x_K$ becomes a random noise from standard Gaussian.

The data generation process learns to reverse the denoising sampling as follows $p_\theta(x_{k-1}|x_k) = \mathcal{N}\big(x_{k-1};\mu_\theta(x_k, k), \sigma_k I \big), 1 \leq k \leq K$,
where $\mu_\theta$ is parameterized with neural network $\theta$ and $\sigma_k^2 = \beta_t$.
In DDPM, the author chooses to parameterize $\eps_\theta$ in replace of $\mu_\theta$: $\mu_\theta(x_k, k) = \frac{1}{\sqrt{\alpha_k}} \biggl( x_k - \frac{\beta_t}{\sqrt{1 - \bar \alpha_t}} \eps_\theta(x_k, k) \biggr)$.
Then the objective of the diffusion model is to predict the noise $\eps_\theta(x_k, k)$ in Eq. \eqref{eq: x_k closed form} from $x_k$.

With the classifier-free guidance \cite{ho2022classifier}, our basline model learns a conditional noise predictor $\eps_\theta(x_k, k, y)$ and a unconditional noise predictor $\eps_\theta(x_k, k, y=\varnothing)$ simultaneously, with the loss function as follows
\begin{align} \label{eq: diffusion model loss}
\mathcal{L}_\text{diff}=
    \E_{(x_0, y), k, \eps, b } \norm{ \eps_\theta\Big(x_k(x_0, \eps), k, (1-b)\cdot y + b \cdot \varnothing \Big) - \eps }^2_2
\end{align}
where $b \sim \text{Bernoulli}(p_\text{uncond})$. 
That is, with the given probability $p_\text{uncond}$, we learn the unconditional $\eps_\theta(x_k, k, y=\varnothing)$ with Eq.\eqref{eq: diffusion model loss} and otherwise learn conditional noise $\eps_\theta(x_k, k, y)$ with the same neural network.

During sampling, we perform the data generation process through the reverse sampling process from $k = K - 1$ to $0$ with the predicted noise $\bar \eps_\theta$:
\begin{align} \label{eq: reverse sample}
    x_{k-1} =  \frac{1}{\sqrt{\alpha_k}} \biggl( x_k - \frac{\beta_t}{\sqrt{1 - \bar \alpha_t}} \bar \eps_\theta(x_k, k, y) \biggr) + \sigma_k z,  \quad 1 \leq k \leq K 
\end{align}
where $\bar \eps_\theta(x, k, y) = (\omega + 1) \eps_\theta(x_k, k, y) - \omega \eps_\theta(x_k, k, y = \varnothing)$ and $\omega$ controls the weight of the conditional generation.
$z \sim \N(0, I)$.
\begin{figure}[!htp]
    \centering
    \includegraphics[width=0.8\textwidth]{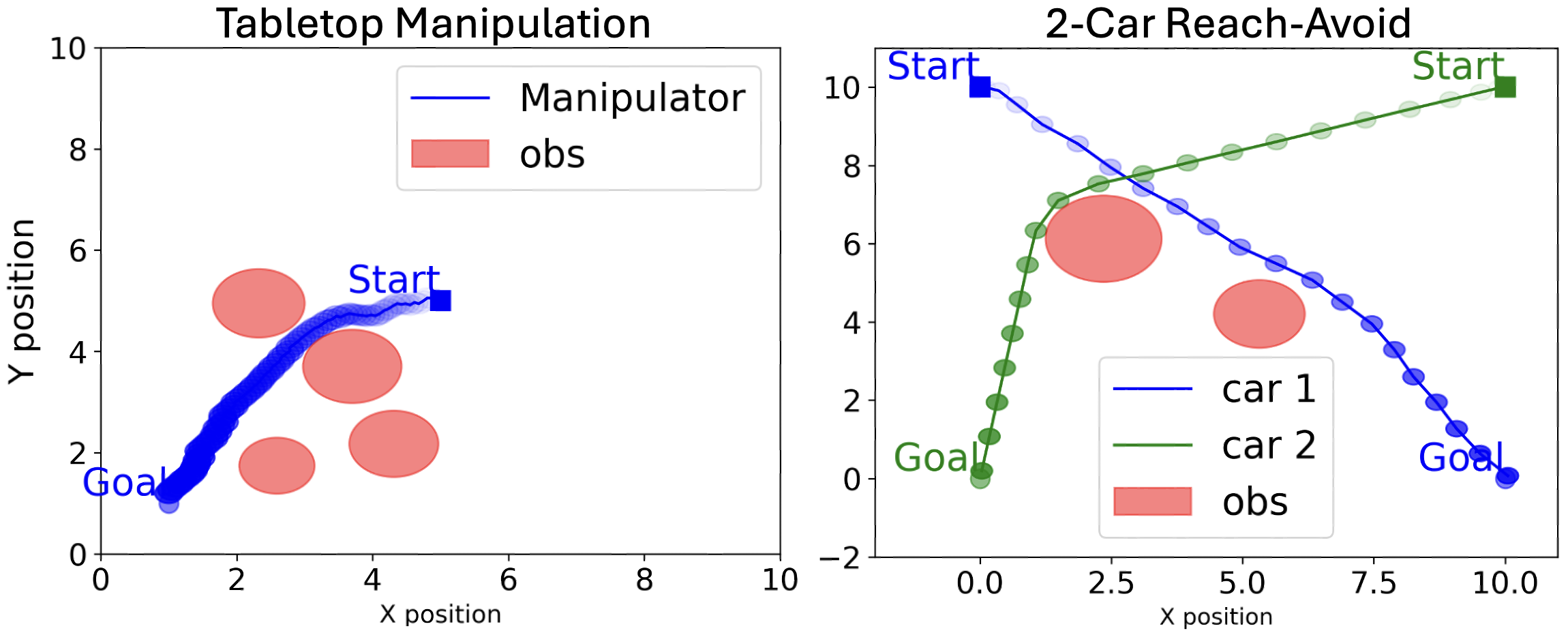}
    \caption{Example trajectory for the tabletop manipulation and 2-car reach-avoid.}
    \vspace{-1em}
    \label{fig: example trajectory}
\end{figure}

\subsection{Constraint-Aware Diffusion Model}

In the previous vanilla model, 
the constraint information is not explicitly incorporated in the training process.
In this section, we present a novel constraint-aware diffusion model that trains with a hybrid objective to sample high-quality solutions while reducing constraint violation.
We first define $V(x;y) \in \mathcal{C}^1(\R^n, \R^k;\R)$ as the sum of the constraint violation value from Eq. \eqref{eq: trajectory optimization prob}:
\begin{align}
    V(x, y) &= \sum_{i=1}^l \max(g_i(x;y), 0) + \sum_{j=1}^m  |h_j(x;y)| \nonumber
\end{align}

Since $V$ evaluates on data $x_k$ and $y$ instead of $\eps_\theta(x_k, k, y)$ from the neural network, we perform a one-step reverse sampling to predict $\tilde x_{k-1}$ from $x_k$ similar to \eqref{eq: reverse sample} but with only conditional noise $\eps_\theta(x_k, k, y)$.
Since the data $\tilde x_k$ is noisy,  we clip $\tilde x_k$ to have the same range of $x_0$.
Now we are able to evaluate the violation $V(\tilde x_{k-1}, y)$, given $x_0, y, k$ and $\eps_\theta$, where the gradient information will be back-propagated through $\eps_\theta$.
We then introduce the violation loss $\mathcal{L}_\text{vio}(x_0, y, k)$:
\begin{align} \label{eq: predicted violation loss}
     \mathcal{L}_\text{vio}(x_0, y, k) &= \mathbb{E}_{\eps, z}\bigg[ V\Big(\tilde x_{k-1}(x_k(x_0, \eps), \eps_\theta, z), y\Big)\bigg]
\end{align}

However, $V(\cdot)$ is not expected to be zero on the noisy data $\tilde x_{k-1}$ or its groundtruth $x_{k-1}$.
Fortunately, we have access to the groundtruth $x_{k-1}$ from the forward diffusion process in Eq. \ref{eq: x_k closed form}.
Thus, we can sample $x_{k-1}$ $N$ times from $\N( \sqrt{\bar{\alpha}_k} x_0, \sqrt{1 - \bar{\alpha}_{k-1}}I)$ and compute the average violation value $\mathcal{\mu}_{\text{vio\_GT}}$ as the groundtruth violation:
\begin{align} \label{eq: groundtruth violation loss mean}
    &\mathcal{\mu}_{\text{vio\_GT}} (x_0, y, k) = \mathbb{E}_{\eps} \bigg[ V\Big( x_{k-1}(x_0, \eps), y\Big)\bigg] 
\end{align}
Finally, we introduce a hybrid loss function from Eq. \ref{eq: diffusion model loss}, \ref{eq: predicted violation loss}, \ref{eq: groundtruth violation loss mean}:
\begin{align} \label{eq: hybrid loss}
    \mathcal{L_{\text{constrained\_diff}}} &= \mathcal{L}_\text{diff} + \lambda \cdot \frac{\mathcal{L}_\text{vio}}{\mathcal{\mu}_{\text{vio\_GT}}}
\end{align}
The violation loss $\mathcal{L}_\text{vio}$ is re-weighted by the groundtruth average of the violation $\mathcal{\mu}_{\text{vio\_GT}}$.
The rationale is that $\mathcal{L}_\text{vio}$ won't be penalized much when the groundtruth violation $\mathcal{\mu}_{\text{vio\_GT}}$ is large, which empirically is the case when the sampling step $k$ is small and the data is less noisy.

\begin{figure}[!htp]
    \centering
    \includegraphics[width=0.9\textwidth]{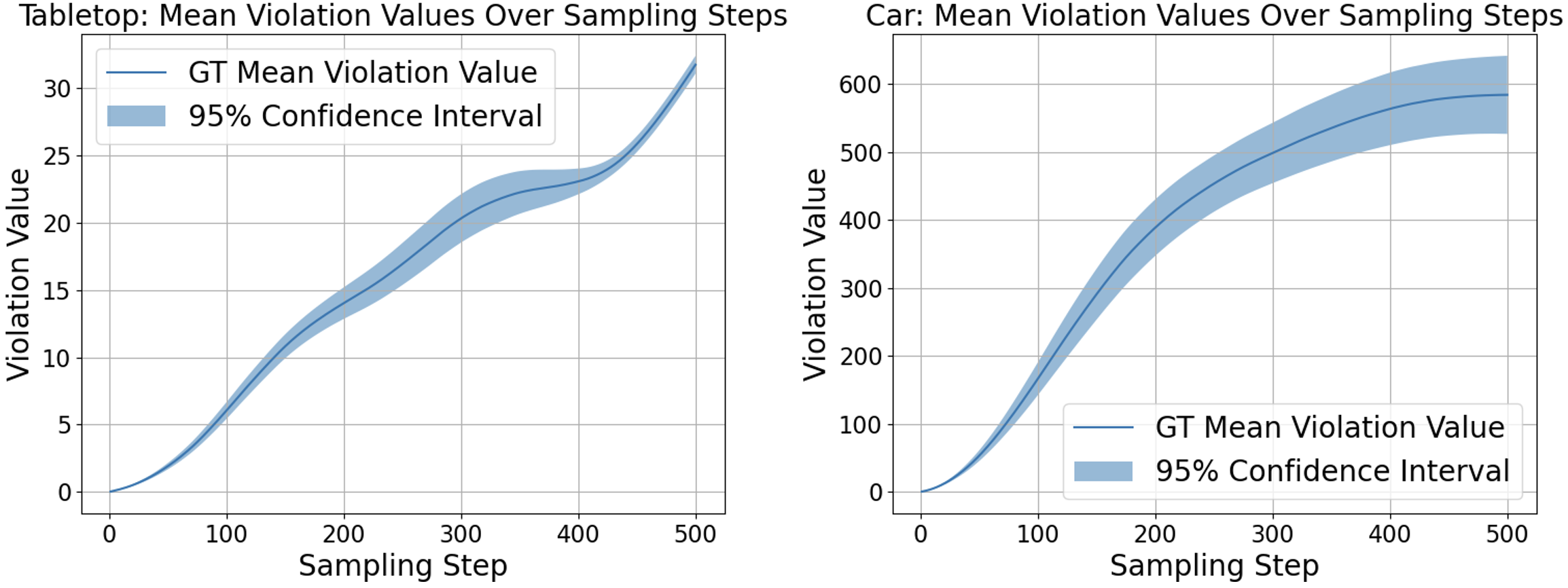}
    \caption{Groundtruth constraint violation. Left: Tabletop Manipulation. Right: Two-car Reach-Avoid}
    \label{fig: tabletop car gt violation}
            \vspace{-1em}
\end{figure}
\vspace{-3em}
\section{Experiment Results and Analysis}

We evaluate our proposed method on two tasks: a tabletop manipulation and a two-car reach-avoid problem.
For both tasks, the constraint-aware and unconstrained diffusion model performs 500 sampling steps both in the forward and reverse process.
Both models use a U-Net architecture \cite{ronneberger2015u} with three hidden layers of 512, 512, and 1024 neurons, and have fully connected layers to encode the conditional information with 2 hidden layers of 256 and 512 neurons.
All models are trained with 3 random seeds, each with 200 epochs using the Adam optimizer \cite{kingma2014adam}.
The training takes 3 to 8 hours for vanilla unconstrained diffusion models and 14 to 55 hours for constraint-aware diffusion models.

\subsection{Tabletop Manipulation}

\subsubsection{Task Setup}
In this task, we aim to minimize the time $t$ for the end factor of the manipulator to reach on the table while avoiding objects, shown in the left of Fig. \ref{fig: example trajectory}. 
We fix the start position in the center of the table.
The varying problem parameter $y = (p_{\text{goal}}, p_{\text{obs}}, r_{\text{obs}})$ 
with the goal $p_{\text{goal}}$ randomly sampled from one of the four corners and the 4 obstacle $p_{\text{obs}}, r_{\text{obs}}$ randomly sampled between the start and goal.
The dynamics of the end factor is modeled as a linear system $\dot p_x = u_x, \dot p_y = u_y$, where $p_x, p_y$ are $x$ and $y$ position.
The controls are $u_x, u_y$.
The trajectory is discretized with 80 timesteps, and thus $x=(t, u_x^1, u_y^1, ..., u_x^{80}, u_y^{80})$.
We use SNOPT to collect 237k locally optimal (feasible) data from 2700 different problems $\mathcal{P}_y$ and test on problems $\mathcal{P}_y$ with unseen $y$ values.

\subsubsection{Results and Analysis}

In the left of Fig. \ref{fig: tabletop car gt violation} we show the increase of groundtruth constraint violation means over 500 sampling steps with 95 \% confidence interval, measured on 128 data with 100 samples on each step.
In Table \ref{table:constraint_violation}, the constraint-aware diffusion model achieves much fewer constraint violations in the first 25\% of the 6k samples and more feasible solution (no violation).
In Table \ref{table:solution_solving_time}, the constraint-aware diffusion model obtains similar locally optimal solutions as the unconstrained diffusion and is only slightly slower when the samples are used to warm-start the numerical solver SNOPT.
It shows that the effort of reducing constraint violation doesn't affect our model's ability to sample solutions close to local optima.

\subsection{Two-Car Reach-Avoid}

\subsubsection{Task Setup}

In this task, we aim to find the trajectory with minimum time $t$ for two cars to reach their own goal while avoiding colliding with each other and the obstacles, shown in the right of Fig. \ref{fig: example trajectory}.
We fix the start and goal position for each car, and set the varying problem parameter $y = (p_{\text{obs}}, r_{\text{obs}})$.
The 2 obstacles $p_{\text{obs}}, r_{\text{obs}}$ are randomly sampled between the start and goal.
The car dynamics are modeled as $\dot p_x = v \cos \theta, \dot p_y = v \sin \theta, \dot v = a, \dot \theta = \omega$, where $p_x, p_y$ are $x, y$ position, $v$ is the velocity, $\theta$ is the orientation.
The controls are acceleration $a$ and angular speed $\omega$.
The trajectory for both car is discretized with 40 timesteps, and thus $x = (t, a_1^1, \omega_1^1, a_2^1, \omega_2^1,..., a_1^{40}, \omega_1^{40}, a_2^{40}, \omega_2^{40})$.
We adopt SNOPT to collect 114000 locally optimal (feasible) solution data from 3k different problem $\mathcal{P}_y$ and test on problems $\mathcal{P}_y$ with unseen $y$ values.



\begin{table}[t]
\caption{Constraint violation values and feasible ratio evaluated before feeding into the solver.}
\label{table:constraint_violation}
\vskip 0.15in
\begin{center}
\begin{small}
\begin{sc}
\begin{tabular}{llccc}
\toprule
Task & Method & Mean($\pm$STD) & 25\%-Quantile & Feasible Ratio \\
\midrule
\multirow{2}{*}{Tabletop} & Constr. Diff.      & 4.80 $\pm$ 5.61 & \textbf{0.04} & \textbf{58.3\textperthousand} \\
                          & Diffusion          & \textbf{4.73 $\pm$ 5.73} & 0.11 & 8.5\textperthousand \\
                          & Uniform            & 32.20 $\pm$ 3.73 & 29.66 & 0\textperthousand\\
\midrule
\multirow{2}{*}{Two-car} & Constr. Diff.      & \textbf{2.72 $\pm$ 20.30} & \textbf{0.54} & \textbf{0.4\textperthousand} \\
                         & Diffusion          & 10.99 $\pm$ 35.59 & 1.98 & 0\textperthousand \\
                         & Uniform            & 545.66 $\pm$ 294.71 & 353.96 & 0\textperthousand \\
\bottomrule
\end{tabular}
\end{sc}
\end{small}
\end{center}
\vskip -0.1in
    \vspace{-1em}
\end{table}
\vspace{-1em}

\begin{table}[t]
\caption{The ratio of locally optimal (implied feasible) solutions obtained with the numerical solver and the corresponding computational time statistics.}
\label{table:solution_solving_time}
\vskip 0.15in
\begin{center}
\begin{small}
\begin{sc}
\begin{tabular}{llccccr}
\toprule
Task & Method   & Ratio & Mean($\pm$STD) & 25\%-Quantile & Median \\
\midrule
\multirow{2}{*}{Tabletop} & Constr. Diff.   & 60\% & 7.63 $\pm$ 16.99 & 1.06 & \textbf{1.36} \\
                          & Diffusion           & 60\% & \textbf{7.51 $\pm$ 17.50} & \textbf{0.73} & 1.37 \\
                          & Uniform             & \textbf{62\%} & 31.65 $\pm$ 24.55 & 14.81 & 22.72 \\
\midrule
\multirow{2}{*}{Two-car} & Constr. Diff.   & 93\% & 19.32 $\pm$ 14.70 & 8.99 & \textbf{15.33} \\
                         & Diffusion           & \textbf{95\%} & \textbf{18.82 $\pm$ 14.33} & \textbf{8.77} & 15.61 \\
                         & Uniform             & 64\% & 46.17 $\pm$ 20.63 & 29.62 & 43.54 \\
\bottomrule
\end{tabular}
\end{sc}
\end{small}
\end{center}
\vskip -0.1in
    \vspace{-1em}
\end{table}

\subsubsection{Results and Analysis}

In the right of Fig. \ref{fig: tabletop car gt violation}, we also notice that the groundtruth violation increases as the data gets noisier in the later sampling steps, and thus the constraint violation loss in Eq. \eqref{eq: hybrid loss} will have smaller weight when $k$ is large.
In Table \ref{table:constraint_violation}, the constraint-aware diffusion has better constraint violation on average and 25\% of the 15k samples and also can produce entirely feasible solutions while unconstrained diffusion and uniform samples cannot produce even one.
In Table \ref{table:solution_solving_time}, the number of locally optimal solutions obtained and computational time for the constraint-aware diffusion model is only slightly worse than the unconstrained diffusion when warm-starting the solver SNOPT using generated samples.
It demonstrates that our proposed model can still generate solutions that are close to local optima while improving the constraint violations of the samples.

\section{Conclusions}

In this paper, we present a novel constraint-aware diffusion model with a hybrid loss function during training for solving trajectory optimization problems.
The proposed model achieves improved constraint satisfaction while generating high-quality solutions, which is demonstrated on tabletop manipulation and 2-car reach-avoid problems.
The limitation is that the current training time of the constraint-aware diffusion model is long due to the complex numerical integration in constraint function and extensive sampling time for obtaining groundtruth violations.
Future work may include more time-efficient ways to incorporate constraint information for the diffusion model such as data augmentation.

%
%

%
%
\bibliographystyle{splncs04}
\bibliography{main.bbl}

\begin{thebibliography}{10}
\providecommand{\url}[1]{\texttt{#1}}
\providecommand{\urlprefix}{URL }
\providecommand{\doi}[1]{https://doi.org/#1}

\bibitem{ajay2022conditional}
Ajay, A., Du, Y., Gupta, A., Tenenbaum, J., Jaakkola, T., Agrawal, P.: Is conditional generative modeling all you need for decision-making? arXiv preprint arXiv:2211.15657  (2022)

\bibitem{boggs1995sequential}
Boggs, P.T., Tolle, J.W.: Sequential quadratic programming. Acta numerica  \textbf{4},  1--51 (1995)

\bibitem{botteghi2023trajectory}
Botteghi, N., Califano, F., Poel, M., Brune, C.: Trajectory generation, control, and safety with denoising diffusion probabilistic models. arXiv preprint arXiv:2306.15512  (2023)

\bibitem{chang2023denoising}
Chang, J., Ryu, H., Kim, J., Yoo, S., Seo, J., Prakash, N., Choi, J., Horowitz, R.: Denoising heat-inspired diffusion with insulators for collision free motion planning. arXiv preprint arXiv:2310.12609  (2023)

\bibitem{chi2023diffusion}
Chi, C., Feng, S., Du, Y., Xu, Z., Cousineau, E., Burchfiel, B., Song, S.: Diffusion policy: Visuomotor policy learning via action diffusion. arXiv preprint arXiv:2303.04137  (2023)

\bibitem{dhariwal2021diffusion}
Dhariwal, P., Nichol, A.: Diffusion models beat gans on image synthesis. Advances in neural information processing systems  \textbf{34},  8780--8794 (2021)

\bibitem{gill2005snopt}
Gill, P.E., Murray, W., Saunders, M.A.: Snopt: An sqp algorithm for large-scale constrained optimization. SIAM review  \textbf{47}(1),  99--131 (2005)

\bibitem{ho2020denoising}
Ho, J., Jain, A., Abbeel, P.: Denoising diffusion probabilistic models. Advances in neural information processing systems  \textbf{33},  6840--6851 (2020)

\bibitem{ho2022classifier}
Ho, J., Salimans, T.: Classifier-free diffusion guidance. arXiv preprint arXiv:2207.12598  (2022)

\bibitem{ho2022video}
Ho, J., Salimans, T., Gritsenko, A., Chan, W., Norouzi, M., Fleet, D.J.: Video diffusion models. Advances in Neural Information Processing Systems  \textbf{35},  8633--8646 (2022)

\bibitem{hoogeboom2022equivariant}
Hoogeboom, E., Satorras, V.G., Vignac, C., Welling, M.: Equivariant diffusion for molecule generation in 3d. In: International conference on machine learning. pp. 8867--8887. PMLR (2022)

\bibitem{janner2022planning}
Janner, M., Du, Y., Tenenbaum, J.B., Levine, S.: Planning with diffusion for flexible behavior synthesis. arXiv preprint arXiv:2205.09991  (2022)

\bibitem{ketata2023diffdock}
Ketata, M.A., Laue, C., Mammadov, R., St{\"a}rk, H., Wu, M., Corso, G., Marquet, C., Barzilay, R., Jaakkola, T.S.: Diffdock-pp: Rigid protein-protein docking with diffusion models. arXiv preprint arXiv:2304.03889  (2023)

\bibitem{kingma2014adam}
Kingma, D.P., Ba, J.: Adam: A method for stochastic optimization. arXiv preprint arXiv:1412.6980  (2014)

\bibitem{krishnamoorthy2023diffusion}
Krishnamoorthy, S., Mashkaria, S.M., Grover, A.: Diffusion models for black-box optimization. arXiv preprint arXiv:2306.07180  (2023)

\bibitem{li2024efficient}
Li, A., Ding, Z., Dieng, A.B., Beeson, R.: Efficient and guaranteed-safe non-convex trajectory optimization with constrained diffusion model. arXiv preprint arXiv:2403.05571  (2024)

\bibitem{maze2023diffusion}
Maz{\'e}, F., Ahmed, F.: Diffusion models beat gans on topology optimization. In: Proceedings of the AAAI conference on artificial intelligence. vol.~37, pp. 9108--9116 (2023)

\bibitem{nichol2021improved}
Nichol, A.Q., Dhariwal, P.: Improved denoising diffusion probabilistic models. In: International conference on machine learning. pp. 8162--8171. PMLR (2021)

\bibitem{power2023sampling}
Power, T., Soltani-Zarrin, R., Iba, S., Berenson, D.: Sampling constrained trajectories using composable diffusion models. In: IROS 2023 Workshop on Differentiable Probabilistic Robotics: Emerging Perspectives on Robot Learning (2023)

\bibitem{rombach2022high}
Rombach, R., Blattmann, A., Lorenz, D., Esser, P., Ommer, B.: High-resolution image synthesis with latent diffusion models. In: Proceedings of the IEEE/CVF conference on computer vision and pattern recognition. pp. 10684--10695 (2022)

\bibitem{ronneberger2015u}
Ronneberger, O., Fischer, P., Brox, T.: U-net: Convolutional networks for biomedical image segmentation. In: Medical image computing and computer-assisted intervention--MICCAI 2015: 18th international conference, Munich, Germany, October 5-9, 2015, proceedings, part III 18. pp. 234--241. Springer (2015)

\bibitem{sohl2015deep}
Sohl-Dickstein, J., Weiss, E., Maheswaranathan, N., Ganguli, S.: Deep unsupervised learning using nonequilibrium thermodynamics. In: International conference on machine learning. pp. 2256--2265. PMLR (2015)

\bibitem{song2020score}
Song, Y., Sohl-Dickstein, J., Kingma, D.P., Kumar, A., Ermon, S., Poole, B.: Score-based generative modeling through stochastic differential equations. arXiv preprint arXiv:2011.13456  (2020)

\bibitem{sun2024difusco}
Sun, Z., Yang, Y.: Difusco: Graph-based diffusion solvers for combinatorial optimization. Advances in Neural Information Processing Systems  \textbf{36} (2024)

\bibitem{trabucco2022design}
Trabucco, B., Geng, X., Kumar, A., Levine, S.: Design-bench: Benchmarks for data-driven offline model-based optimization. In: International Conference on Machine Learning. pp. 21658--21676. PMLR (2022)

\bibitem{yang2023compositional}
Yang, Z., Mao, J., Du, Y., Wu, J., Tenenbaum, J.B., Lozano-P{\'e}rez, T., Kaelbling, L.P.: Compositional diffusion-based continuous constraint solvers. arXiv preprint arXiv:2309.00966  (2023)

\end{thebibliography}

\end{document}